\documentclass[10pt,twocolumn,letterpaper]{article}

\usepackage[pagenumbers]{cvpr} 
\usepackage{url}
\usepackage{graphicx}
\usepackage{wrapfig}
\usepackage{subcaption}
\usepackage{booktabs}       
\usepackage{multirow} 
\usepackage{amssymb}
\usepackage{pifont}
\usepackage{algorithm}
\usepackage{algorithmic}
\usepackage{pifont}
\usepackage{bm}
\usepackage{cuted}      
\usepackage{caption}    

\usepackage{color} 

\usepackage{xcolor}
\usepackage{colortbl,booktabs}
\definecolor{cvprblue}{rgb}{0.21,0.49,0.74}
\usepackage[pagebackref,breaklinks,colorlinks,allcolors=cvprblue]{hyperref}

\newcommand{\cmark}{\textcolor{green!60!black}{\ding{51}}}  
\newcommand{\xmark}{\textcolor{red!70!black}{\ding{55}}}    

\definecolor{citecolor}{RGB}{17,80,197}
\hypersetup{
    colorlinks=true,
    linkcolor=red,
    citecolor=citecolor,
    filecolor=magenta,      
    urlcolor=magenta,
}
\def\paperID{2225} 
\def\confName{CVPR}
\def\confYear{2026}

\title{Rectified SpaAttn: Revisiting Attention Sparsity for Efficient Video Generation}
\vspace{-5cm}
\author{Xuewen Liu$^{1,2}$, \; Zhikai Li$^{1}$\thanks{Corresponding author: \{zhikai.li, qingyi.gu\}@ia.ac.cn.}, \; Jing Zhang$^{1,2}$, \; Mengjuan Chen$^{1}$, \; Qingyi Gu$^{1}$\footnotemark[1] \\
$^1$Institute of Automation, Chinese Academy of Sciences\\
$^2$School of Artificial Intelligence, University of Chinese Academy of Sciences\\
{\tt\small \{liuxuewen2023, zhikai.li, qingyi.gu\}@ia.ac.cn}
}

\begin{document}

\vspace{-5cm}
\maketitle
\vspace{-10cm}

\begin{strip}
    \centering
    \includegraphics[width=0.98\textwidth]{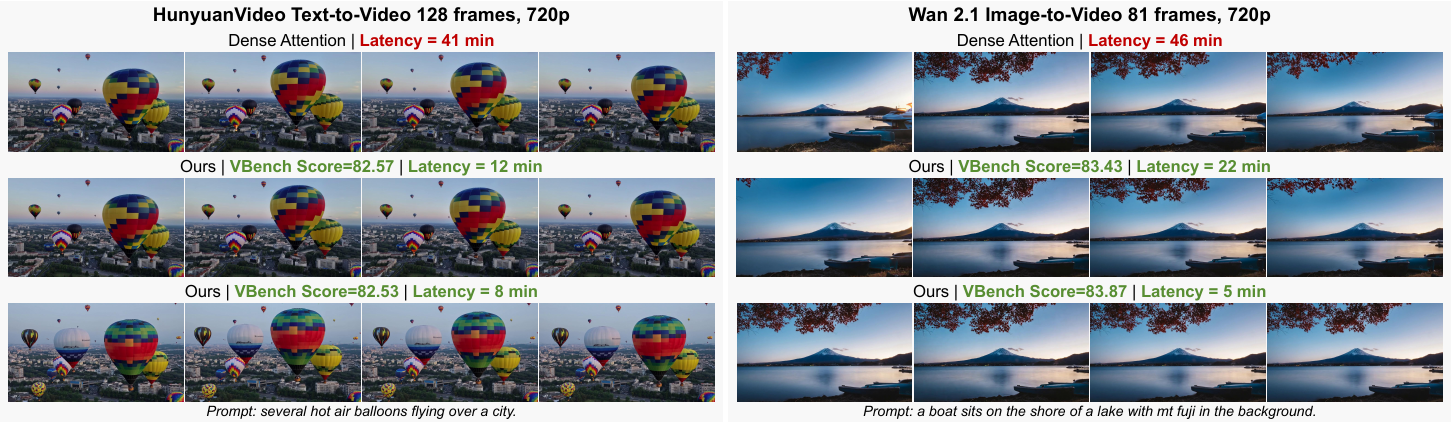}
    \vspace{-0.2cm}
    \captionof{figure}{Rectified SpaAttn achieves significant speedup while maintaining high quality, making video generation more efficient.}
    \label{fig:motivation}
\end{strip}

\begin{abstract}
Diffusion Transformers dominate video generation, but the quadratic complexity of attention computation introduces substantial latency, hindering real-world applications.
Attention sparsity reduces computational costs by focusing on critical tokens while ignoring non-critical tokens. 
However, existing methods suffer from severe performance degradation, especially under high sparsity.
In this paper, we revisit attention sparsity and show, through theoretical analysis, that existing methods induce systematic biases in attention allocation:
(1) excessive focus on critical tokens amplifies their attention weights; (2) complete neglect of non-critical tokens causes the loss of relevant attention weights.
To address these issues, we propose Rectified SpaAttn, which rectifies sparse attention allocation with implicit full attention reference, thereby enhancing the alignment between sparse and full attention maps.
Specifically, we implicitly capture the full-attention distribution of critical and non-critical tokens by pooled query-key interactions, without explicit computation, which is then used to rectify sparse attention allocation: 
(1) for critical tokens, we show that their bias is proportional to the sparse attention weights, with the ratio governed by the amplified weights. Accordingly, we propose Isolated-Pooling Attention Reallocation, which calculates accurate rectification factors by reallocating multimodal pooled weights.
(2) for non-critical tokens, recovering attention weights from the pooled query-key yields attention gains but also introduces pooling errors. Therefore, we propose Gain-Aware Pooling Rectification, which ensures that the rectified gain consistently surpasses the induced error.
Moreover, we customize and integrate the Rectified SpaAttn kernel using Triton, achieving up to 3.33× and 2.08× speedups on HunyuanVideo and Wan 2.1, respectively, while maintaining high generation quality.
We release Rectified SpaAttn as open-source at \href{https://github.com/BienLuky/Rectified-SpaAttn}{https://github.com/BienLuky/Rectified-SpaAttn}.

\end{abstract}    
\vspace{-0.5cm}
\section{Introduction}
\vspace{-0.2cm}
Diffusion Transformers (DiTs)~\cite{peebles2023scalable} have achieved remarkable success in video generation tasks~\cite{kong2024hunyuanvideo,wan2025wan}. However, the inference efficiency of DiTs remains a major bottleneck, primarily due to the quadratic complexity of attention computation~\cite{vaswani2017attention}. 
As an example, generating a 128-frame 720p video with HunyuanVideo~\cite{kong2024hunyuanvideo} on an NVIDIA H100 PCIe requires approximately 41 minutes end-to-end, where attention mechanisms dominate 32 minutes of the latency even with FlashAttention~\cite{dao2023flashattention} optimization.

To reduce the computational complexity of attention mechanisms, attention sparsity~\cite{tang2024quest,xiao2024efficient} restricts each query token to interact with only a small subset of critical key-value tokens, thereby omitting most non-critical computations.
Typically, the importance of a token is determined by its full-attention weight: tokens with larger weights are regarded as critical, while those with smaller weights are considered non-critical. To support block-wise computation optimizations such as FlashAttention~\cite{dao2023flashattention}, this identification is performed at the block level.
In addition, since visual tokens within a block exhibit strong homogeneity, existing methods~\cite{xia2025training,xi2025sparse,zhang2025training,zhang2025fast} evaluate visual block importance using pooled attention weights derived from uniformly pooled query-key interactions.
It significantly reduces the overhead of evaluation.
After evaluation, sparse mask~\cite{zhang2025training} identifies critical tokens, guiding the subsequent sparse computation.

Although existing methods~\cite{xi2025sparse,ren2025grouping,li2025radial,yang2025sparse,zhang2025training} remain effective at low sparsity ratios, their performance degrades drastically at higher levels. For example, The Vision Reward~\cite{xu2024visionreward} of Jenga~\cite{zhang2025training} drops to 0.0585 under 90\% sparsity ratio.
To investigate the underlying cause, 
we revisit attention sparsity and reveal that existing methods induce systematic biases in attention allocation, as shown in Fig.~\ref{fig:motivation}.

\noindent \textbf{(1) Allocation bias of critical tokens:} Since attention sparsity involves only critical tokens in the softmax normalization, their attention weights are amplified relative to full attention.
For instance, a critical token with a weight of 0.8 under full attention may rise to 1.0 under sparse attention.
As a result, the attention outputs of critical tokens deviate from those of full attention, and the bias becomes more pronounced as the sparsity ratio increases.

\noindent \textbf{(2) Allocation bias of non-critical tokens:} Attention sparsity completely omits computations for non-critical tokens, causing their attention weights to be discarded. 
For instance, non-critical tokens with attention weights of 0.2 under full attention will become 0.0 under sparse attention. 
This results in information loss compared to full attention, and the bias also further increases as sparsity ratio grows.

\begin{figure*}[t]
    \vspace{-0.2cm}
    \centering
    \begin{subfigure}[b]{0.55\textwidth}
        \centering
        \includegraphics[width=\textwidth]{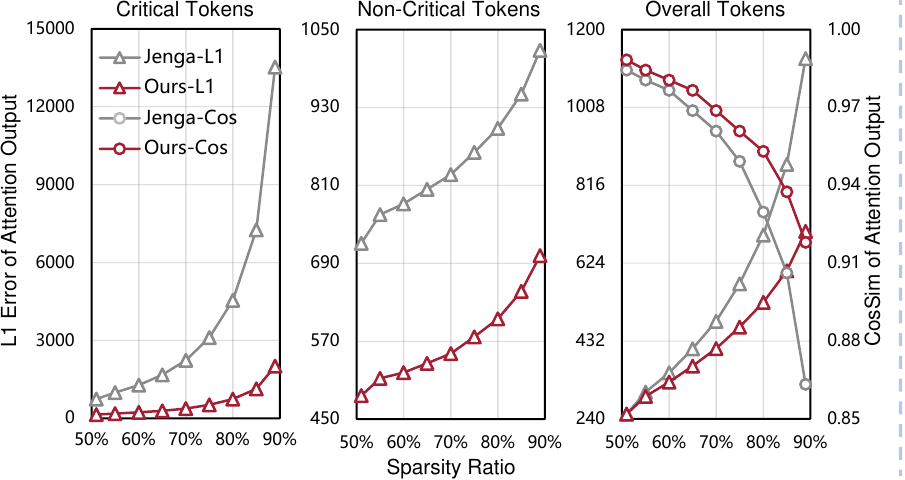}
        \vspace{-0.5cm}
        \caption{Allocation biases result in attention output error}
        \label{fig:L1_CosSim}
    \end{subfigure}
    \hfill
    \begin{subfigure}[b]{0.44\textwidth}
        \centering
        \includegraphics[width=\textwidth]{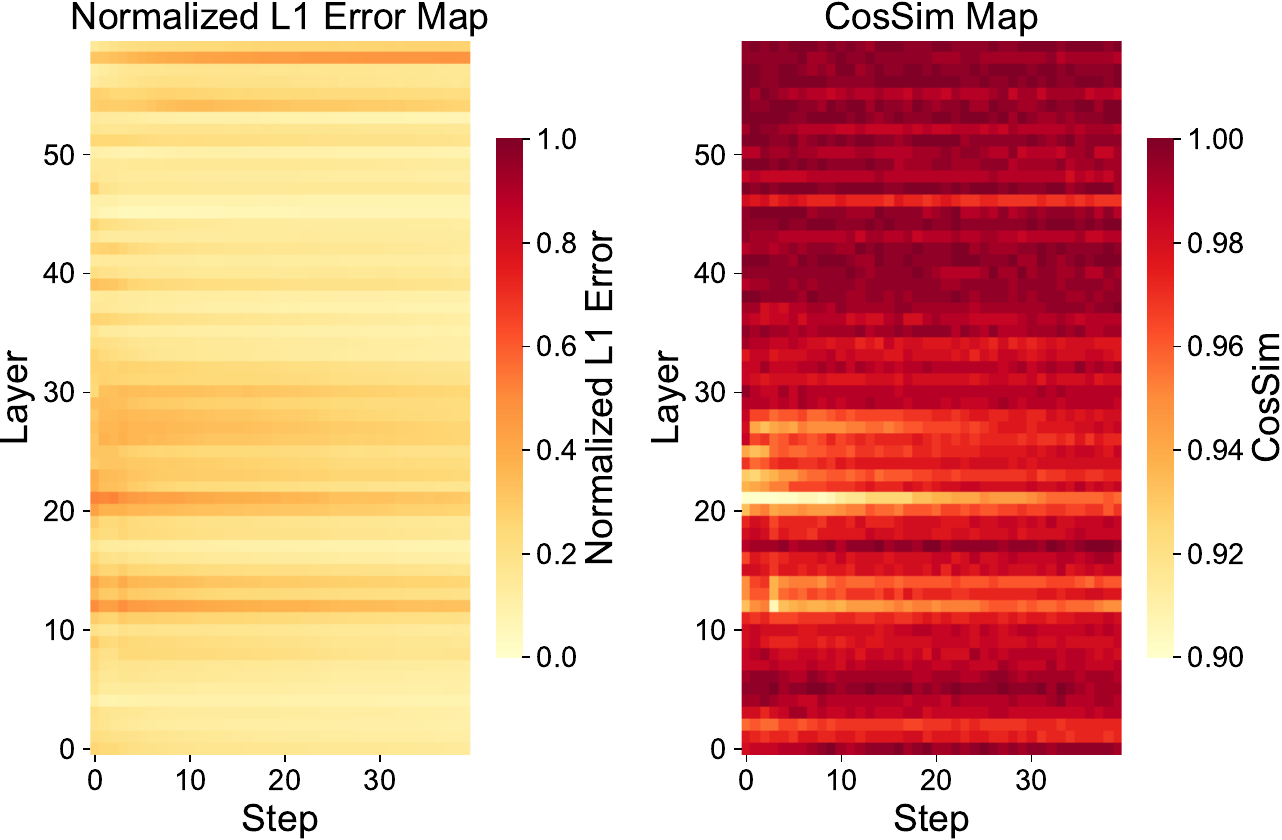}
        \vspace{-0.5cm}
        \caption{Implicit full attention aligned with true attention}
        \label{fig:qkv_compare}
    \end{subfigure}
    \vspace{-0.3cm}
    \caption{\textbf{Motivation.} (a) Attention sparsity introduces allocation biases that cause large output errors, which intensify as sparsity increases. (b) Implicit full attention obtained by IPAR exhibits strong alignment with the true full attention in both magnitude and distribution, as evidenced by lower normalized L1 error and higher cosine similarity across timesteps and layers under 80\% sparsity. The normalized L1 error is computed as the L1 difference normalized by the magnitude of the true attention. Data come from HunyuanVideo.}
    \label{fig:qkv_maps}
    \vspace{-0.3cm}
\end{figure*}

To address the above issues, we propose \textbf{Rectified SpaAttn}, a training-free framework that rectifies sparse attention allocation to enhance the alignment between sparse and full attention maps.
Our method is motivated by a key observation: \textit{Pooled attention weights derived from uniformly pooled query-key interactions can implicitly capture the full-attention distribution of critical and non-critical tokens.} (as shown in Fig.~\ref{fig:qkv_compare}).
Therefore, we leverage them as implicit full attention to guide the rectification: 

\noindent \textbf{(1) For critical tokens}, we demonstrate that their bias is proportional to the attention weights, with the ratio determined by the amplified weights.
Thus, the bias can be rectified by the full-attention weights.
Based on this, we calculate rectified factor using the pooled attention weights, avoiding the costly explicit full attention.
However, since text tokens lack intra-block homogeneity, directly pooling them causes large errors.
To address this, we propose Isolated-Pooling Attention Reallocation (IPAR), which isolates text tokens during pooling and reallocates weights afterward to obtain accurate pooled attention weights.

\noindent \textbf{(2) For non-critical tokens}, we recover attention weights using the pooled attention map.
However, since the rectification relies on block-level pooled weights to approximate token-level weights, it yields attention gains but also introduce pooling errors.
To this end, we propose Gain-Aware Pooling Rectification (GAPR), which estimates the attention gains and pooling errors, and performs rectification only when the gain surpasses the error, thereby ensuring stable performance improvement.


\begin{figure}[!t]
    \centering
    \includegraphics[width=0.45\textwidth]{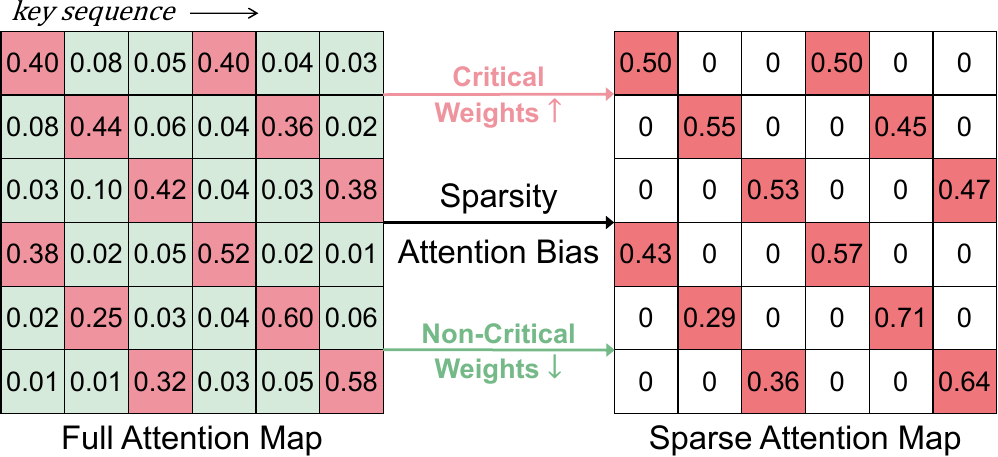}
    \vspace{-0.3cm}
    \caption{Sparsity results in systematic attention biases.}
    \label{fig:motivation}
    \vspace{-0.7cm}
\end{figure}

We evaluate our method on representative generative models, including HunyuanVideo-T2V~\cite{kong2024hunyuanvideo}, Wan2.1-T2V/\,I2V~\cite{wan2025wan}, and Flux.1-dev~\cite{blackforest2024flux}.
Results demonstrate that Rectified SpaAttn outperforms existing methods~\cite{xi2025sparse,yang2025sparse,li2025radial,ren2025grouping,zhang2025training} across various sparsity ratios, particularly under higher sparsity.
Specifically, Rectified SpaAttn achieves up to a 3.33× end-to-end speedup under 90\% sparsity ratio, while maintaining high visual quality with a VBench Score~\cite{huang2024vbench} of up to 82.57 on HunyuanVideo.
By further combining with caching techniques~\cite{liu2025timestep}, our method achieves 5.24×, 8.97×, and 4.15× speedups on HunyuanVideo-T2V, Wan2.1-I2V, and Flux.1-dev, respectively.
Overall, our key contributions are as follows:
\begin{itemize}
    \item We revisit attention sparsity, and reveal the limitations of existing methods as systematic biases in attention allocation: excessive focus on critical tokens amplifies their attention weights, while complete neglect of non-critical tokens causes the loss of relevant attention weights.
    \item We propose Rectified SpaAttn, which leverages implicit full attention to guide the rectification of allocation biases, aligning sparse attention with its full counterpart. We introduce IPAR and GAPR to ensure accurate rectification for critical and non-critical tokens, respectively.
    \item Rectified SpaAttn pushes the sparsity limit while preserving high generation quality. 
    We further optimize it with Triton and package it as a plug-and-play attention module for efficient and scalable deployment.
\end{itemize}

\section{Related Work}
\vspace{-0.1cm}
\textbf{Efficient Video Generation.}
The high inference cost of video generation~\cite{li2025k} remains a major bottleneck, severely constraining their applicability in real-world scenarios.
To reduce computation, timestep compression~\cite{geng2025mean} shorten the trajectory by rectifying generation objectives.
Model compression include model pruning~\cite{castells2024ld} to reduce parameter scale, model quantization~\cite{liu2024eda,liu2025dilatequant} to lower parameter bit-width, and model distillation~\cite{meng2023distillation} to construct lightweight models.
Additionally, model caching~\cite{ma2024deepcache,liu2025cachequant,liu2025timestep} leverage the high similarity between timesteps, improving efficiency through feature reuse.
Although these methods have achieved remarkable success, they are mostly applied to low-resolution image or short video generation.
With the growing demand for long video generation, models are required to process 50K$\sim$200K tokens.
The quadratic complexity of attention thus causes a drastic rise in computation, becoming the main bottleneck of inference. 

\noindent \textbf{Attention Sparsity.}
Previous studies~\cite{zhang2023h2o,xiao2024efficient} have shown that attention in transformers exhibit nature sparsity.
Building on this, several works~\cite{tang2024quest,jiang2024minference} retain only the interactions between query tokens and critical key-value tokens to reduce attention computation, known as attention sparsity.
SVG~\cite{xi2025sparse} explores it for DiTs by designing spatial and temporal sparse mask.
AdaSpa~\cite{xia2025training} dynamically evaluates token importance and adaptively constructs blockified sparse mask.
STA~\cite{zhang2025fast} and GRAT~\cite{ren2025grouping} leverage the locality of attention by maintaining fixed tiled attention windows.
RadialAttention~\cite{li2025radial} exploits the decay of attention and design a radial sparse mask.
SVG2~\cite{yang2025sparse} applies k-means clustering to reorder tokens and defines block-wise sparse masks.
Jenga~\cite{zhang2025training} reorders token sequences using space-filling curves to enhance intra-block coherence.
Despite notable progress, most methods reach only around 75\% sparsity.
At higher sparsity, the performance of them degrades sharply, making it difficult to balance speedup and quality.

\section{Motivation}

\subsection{Systematic Biases in Attention Allocation}\label{sec:3.1}
In DiTs, attention integrates video and text tokens into a unified sequence~\cite{kong2024hunyuanvideo}.
Specifically, video queries $\bm Q_{v}\in \mathbb{R}^{T_{v} \times d}$ handle video–video and video–text attention, while text queries $\bm Q_{t}\in \mathbb{R}^{T_{t}\times d}$ handle text–video and text–text attention, where $T_v$ and $T_t$ denote the token lengths of video and text, respectively, and $d$ is the feature dimension.
Since $T_v \gg T_t$ and video tokens exhibit redundancy, existing methods~\cite{xi2025sparse,xia2025training,zhang2025fast,ren2025grouping,li2025radial,yang2025sparse,zhang2025training} apply sparsity only to $\bm Q_{v}$ while retaining full attention for $\bm Q_{t}$.
For simplicity, we use $\bm Q$ to denote $\bm Q_{v}$ and focus on its attention.
Additionally, to reduce frequent memory access, attention computation is typically performed in block level\cite{dao2023flashattention}, where consecutive $\bm Q$ and $\bm {KV}$ are aggregated into a block, denoted as:
\begin{align}
    \bm Q &= [\bm Q_1, \bm Q_2, \dots, \bm Q_N], \; \bm Q_n \in \mathbb{R}^{B^q \times d}\\
    \bm K &= [\bm K_1, \bm K_2, \dots, \bm K_M], \; \bm K_m \in \mathbb{R}^{B^k \times d} \\
    \bm V &= [\bm V_1, \bm V_2, \dots, \bm V_M], \; \bm V_m \in \mathbb{R}^{B^k \times d} 
\end{align}
Here, $B^q$ and $B^k$ denotes the block sizes of query and key, respectively. In our experiments, we set $B=B^q=B^k=128$. $N=\frac{T_{v}}{B}$ and $M=\frac{T_{v}+T_{t}}{B}$ represent the number of blocks for $\bm Q$ and $\bm {KV}$, respectively. 
The full attention weight for the $\bm Q_n$ can be formulated as:
\begin{align}\label{eq:Wn}
    \bm S_n=\frac{\bm Q_n \bm K^T}{\sqrt{d}}, \; \bm A_n = \text{softmax}(\bm S_n)
\end{align}

\begin{figure*}[!t]
    \centering
    \vspace{-0.3cm}
    \includegraphics[width=1.0\textwidth]{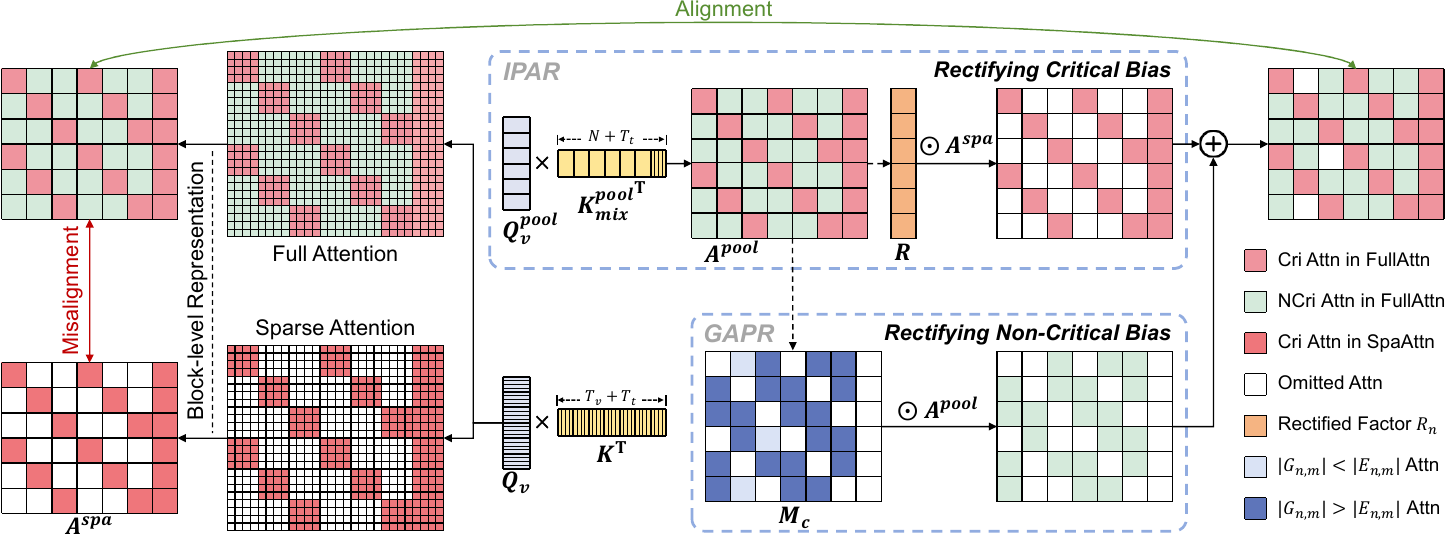}
    \caption{\textbf{Overview of Rectified SpaAttn.} Sparse attention is computed at the block level, exhibiting significant misalignment with full attention. IPAR isolates text tokens during pooling and reallocates pooled weights to obtain distribution-aligned implicit full attention, rectifying biases of critical blocks. GAPR estimates rectified gains and pooling errors to generate compensation mask, applying implicit attention compensation to non-critical blocks. The rectified sparse attention achieves strong alignment with the full attention.}
    \label{fig:overview}
    \vspace{-0.5cm}
\end{figure*}

Attention sparsity reduces computational cost by limiting the interaction number of $\bm {KV}$ tokens.
To align with block-wise computation, sparsity is also applied at the block level: if a block is identified as critical, all $\bm {KV}$ tokens within it are preserved; otherwise, the entire block is discarded, as illustrated in Fig.~\ref{fig:overview}.
Existing methods~\cite{xi2025sparse,li2025radial,zhang2025training} evaluate the importance of block to generate sparse masks $\widehat{\bm M} \in \{0,1\}^{N \times M}$, where
$\widehat{M}_{n,m}=1$ indicates that $\bm Q_n$ is permitted to interact with $\bm K_m$ and $\bm V_m$, and $\widehat{M}_{n,m}=0$ otherwise.
Therefore, the sparse attention weight for the $\bm Q_n$ is formulated as:
\begin{align}\label{eq:Wnspa}
    \bm S_n^{spa} = \frac{\bm Q_n \bm K^T\odot \widehat{\bm M}_n}{\sqrt{d}}, \; \bm A_n^{spa} = \text{softmax}(\bm S_n^{spa})
\end{align}

According to Eq.~\ref{eq:Wn} and Eq.~\ref{eq:Wnspa}, our theoretical analysis reveals that attention sparsity induces systematic allocation biases, leading to large output errors, as shown in Fig.~\ref{fig:L1_CosSim}.

\noindent \textbf{(1) For critical tokens}, assuming the $\bm K_{m-1}$, their attention weights for $\bm Q_n$ in full and sparse attention are donates as:
\begin{equation}\label{eq:Wnmspacri}
\begin{aligned}
    \bm A_{n,m-1}&=\frac{\text{exp}(\bm S_{n,m-1})}{{\textstyle \sum_{m=1}^{M}}\text{exp}(\bm S_{n,m})}\\
    \bm A_{n,m-1}^{spa}&=\frac{\text{exp}(\bm S_{n,m-1})}{{\textstyle \sum_{m=1}^{M}}\text{exp}(\bm S_{n,m}) \cdot \widehat{M}_{n,m}}
\end{aligned} 
\end{equation}
As can be seen, due to the reduced number of elements involved in softmax normalization, $|\widehat{\bm M}_n|<M$, $|\widehat{\bm M}_n|$ denoting the number of preserved $\bm K$ blocks for $\bm Q_n$, their attention weights are amplified compared to those in full attention, $\bm A_{n,m-1}^{spa}>\bm A_{n,m-1}$.
This is regarded as an excessive focus on critical tokens, which introduces attention bias.

\noindent \textbf{(2) For non-critical tokens}, assuming the $\bm K_{m}$, their attention weights for $\bm Q_n$ are formulated as:
\begin{equation}\label{eq:Wnmspancri}
\begin{aligned}
    \bm A_{n,m}&=\frac{\text{exp}(\bm S_{n,m})}{{\textstyle \sum_{m=1}^{M}}\text{exp}(\bm S_{n,m})}\\
    \bm A_{n,m}^{spa}&=\frac{0}{{\textstyle \sum_{m=1}^{M}}\text{exp}(\bm S_{n,m}) \cdot \widehat{M}_{n,m}}
\end{aligned}  
\end{equation}
Due to the complete omission of computation, their attention weights are discarded compared to those in full attention, $\bm A_{n,m}^{spa}=0$.
This results in the loss of their attention, thereby also introducing attention bias.

Clearly, these biases intensify as the sparsity increases.
Although existing methods strive to design sparse masks, the performance degrades significantly at high sparsity.

\subsection{Implicit Full Attention}\label{sec:3.2}
Based on the analysis in Sec~\ref{sec:3.1}, we explore \textit{how to rectify the systematic biases to enhance the alignment between sparse and full attention maps}.
Evidently, the rectification can be achieved by leveraging the full attention weights.
However, explicit full attention incurs significant computational overhead, which contradicts the purpose of attention sparsity and undermines its acceleration benefits.

Previous methods~\cite{jiang2024minference,zhang2025training} uniformly pooled the visual $\bm Q \bm K_v$ to compute relative pooled attention weights $\bm A_r^{pool}\in \mathbb{R}^{N \times N}$, which are used to evaluate the importance of visual blocks.
In contrast, we observe that the pooled attention weights $\bm A^{pool}\in \mathbb{R}^{N \times M}$, derived from uniformly pooled the complete $\bm {QK}$ interactions, inherently possess the capacity to implicitly capture the full-attention distribution of critical and non-critical tokens.
Specifically, they exhibit strong consistency with full-attention weights in both magnitude and distribution, as shown in Fig.~\ref{fig:qkv_compare}.
Therefore, it can not only evaluate block importance but also guide bias rectification.
The formulation is expressed as:
\begin{equation}
\begin{aligned}
    Q_n^{pool}&=\frac{1}{B}\sum_{i\in B_n^q} Q_i \\
    \bm K^{pool}&=[\frac{1}{B}\sum_{j\in B_1^k} K_j, \frac{1}{B}\sum_{j\in B_2^k} K_j, \dots, \frac{1}{B}\sum_{j\in B_M^k} K_j] \\
    \bm S_n^{pool}&= \frac{Q_n^{pool} {\bm K^{pool}}^T}{\sqrt{d}}, \; \bm A_n^{pool} = \text{softmax}(\bm S_n^{pool})
\end{aligned}    
\end{equation}
where $B_n^q$ and $B_m^k$ represent the index sets within query and key blocks, respectively.
$Q_n^{pool}$ and $\bm K^{pool}$ denote the uniformly pooled representations of $\bm Q_n$ and $\bm K$.
Note that directly pooling text tokens in $\bm{K}$ leads to information loss, which we address in Sec.~\ref{sec:4.1}.
As the computation is performed on the pooled $\bm {QK}$ blocks, the computational complexity is significantly reduced.
Moreover, since $T_v \gg T_t$, it introduces only a negligible overhead compared with $\bm A_r^{pool}$.
Given its advantages in accuracy and efficiency, we regard the pooled attention weights as implicit full attention to guide the rectification of systematic biases.

\section{Rectified SpaAttn}
\subsection{Rectifying the Bias of Critical Tokens}
For critical tokens, assuming the $\bm K_m$, their sparse attention weights satisfy $\bm A^{spa}_{n,m}>\bm A_{n,m}$ as analyzed in Sec~\ref{sec:3.1}, indicating an attention amplification bias.
To rectify this bias, we aim to align $\bm A^{spa}_{n,m}$ with $\bm A_{n,m}$.
According to Eq.~\ref{eq:Wnmspacri}, we find that the sparse attention weights are proportional to the full-attention weights.
Therefore, we introduce a rectified factor $\bm R\in \mathbb{R}^{N \times M}$, which can be formulated as:
\begin{equation}
\begin{aligned}
    R_{n,m} &= \frac{\bm A_{n,m}}{\bm A^{spa}_{n,m}}
    = \frac{{\textstyle \sum_{m=1}^{M}}\text{exp}(\bm S_{n,m}) \cdot \widehat{M}_{n,m}}{{\textstyle \sum_{m=1}^{M}}\text{exp}(\bm S_{n,m})} \\
    &= {\textstyle \sum_{m\in \{m|\widehat{M}_{n,m}=1\}}}\bm A_{n,m}
\end{aligned}  
\end{equation}
It can be seen that $R_{n,m} = R_{n,m'} = R_n, \; \forall \; m,m' \in \{1,\dots,M\}$, and thus it can be simplified as $\bm R\in \mathbb{R}^{N}$. Furthermore, $R_n$ essentially represents the total weights of the critical tokens in the original full-attention.
Considering the high cost of explicitly computing full attention, we implicitly estimate $\bm R$ using pooled attention weights.

\begin{figure}[h]
    \centering
    \begin{subfigure}[b]{0.12\textwidth}
        \centering
        \includegraphics[width=\textwidth]{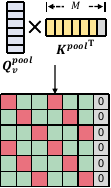}
    \end{subfigure}
    \hspace{0.01\textwidth}
    \begin{subfigure}[b]{0.33\textwidth}
        \centering
        \includegraphics[width=\textwidth]{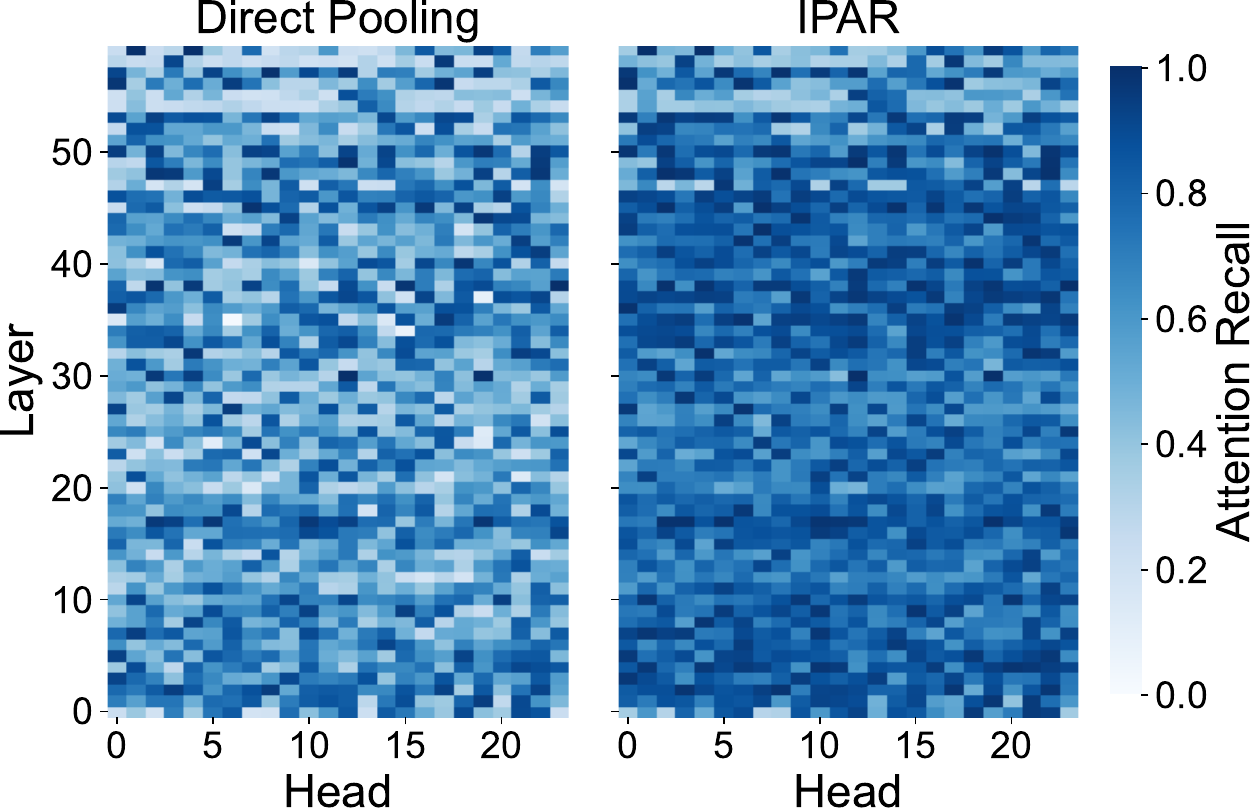}
    \end{subfigure}
    
    \caption{Left: Attention maps from direct pooling. Right: Attention recall of HunyuanVideo under 80\% sparsity.}
    \label{fig:directpool}
    \vspace{-0.3cm}
\end{figure}

However, unlike video tokens, text tokens lack intra-block homogeneity; therefore, applying uniform pooling to them results in attention loss.
Moreover, prior studies~\cite{chen2025sparse,zhang2025training} have shown that text tokens occupy high attention weights in certain attention heads.
Therefore, directly pooling disrupts the attention consistency, failing to accurately approximate the full-attention, as shown in Fig.~\ref{fig:directpool}.

To address this issue, we propose \textbf{Isolated-Pooling Attention Reallocation (IPAR)}. \label{sec:4.1}
Specifically, we isolate text tokens $\bm K_{t}\in \mathbb{R}^{T_{t} \times d}$, and perform block-level pooling only on video tokens $\bm K_{v}\in \mathbb{R}^{T_{v} \times d}$ to achieve $\bm K_{v}^{pool}\in \mathbb{R}^{N \times d}$, yielding a mixed-granularity key sequence $\bm K_{mix}^{pool}\in \mathbb{R}^{(N+T_{t}) \times d}$, on which the pooled attention weights $\bm A_{mix}^{pool}\in \mathbb{R}^{N \times (N+T_{t})}$ are computed:
\begin{align}
    \bm A_{mix}^{pool} = \text{softmax}(\frac{\bm Q^{pool} {\bm K_{mix}^{pool}}^T}{\sqrt{d}})
\end{align}
where the attention weights for video and text tokens are denoted as $\bm A_{v}^{pool} \in \mathbb{R}^{N \times N}$ and $\bm A_{t} \in \mathbb{R}^{N \times T_{t}}$, respectively.
Due to the different computation granularity between $\bm A_{v}^{pool}$ and $\bm A_{t}$, block-level for the former and token-level for the latter, $\bm A_{mix}^{pool}$ exhibits an imbalance in cross-modal attention strength.
To this end, we further propose a weight reallocation strategy.
We reweight the $\bm A_{v}^{pool}$ by block size $B$, ensuring that the weights of each block aligns with the token-level granularity.
Then, we perform re-normalization:
\begin{equation}
\begin{aligned}
    \hat{\bm A}_{v}^{pool} = \frac{\bm A_{v}^{pool} \cdot B}{\bm A_{v}^{pool} \cdot B + \bm A_{t} }, \,
    \hat{\bm A}_{t} = \frac{\bm A_{t}}{\bm A_{v}^{pool} \cdot B + \bm A_{t} } 
\end{aligned}   
\end{equation}
Here, $\hat{\bm A}_{v}^{pool}$ and $\hat{\bm A}_{t}$ denote the token-level weights after reallocation. We further aggregate $\hat{\bm A}_{t}$ by blocks to obtain the block-wise attention $\hat{\bm A}_{t}^{pool}\in \mathbb{R}^{N \times \frac{T_{t}}{B}}$, and concatenate it with $\hat{\bm A}_{v}^{pool}$ to form the final implicit full attention $\bm A^{pool}\in \mathbb{R}^{N \times M}$. 
IPAR produces cross-modally consistent pooled attention weights that accurately approximate full attention, thereby ensuring reliable computation of $R_n$ to achieve rectified sparse attention ${\bm A^{spa}_{n,m}}'$ of the critical tokens, as shown in Fig.~\ref{fig:overview}.
\begin{equation}
\begin{aligned}
    R_n &= {\textstyle \sum_{m\in \{m |\widehat{M}_{n,m}=1\}}}\bm A_{n,m}^{pool} \\
    &{\bm A^{spa}_{n,m}}'=\bm A_{n,m}^{spa} \cdot R_n \approx \bm A_{n,m}
\end{aligned}
\end{equation}

\subsection{Rectifying the Bias of Non-Critical Tokens}
In sparse attention, non-critical tokens are excluded from computation, lossing their attention weights, as shown in Eq.~\ref{eq:Wnmspancri}.
To rectify this bias, we compensate $\bm A_{n,m}^{spa}$ using the pooled attention weights $A_{n,m}^{pool}$, aligning it with $\bm A_{n,m}$.
To evaluate $A_{n,m}^{pool} \approx \bm A_{n,m}$ within the same granularity space, we remap the block-level weights $A_{n,m}^{pool}\in \mathbb{R}^{1}$ to the token-level weights $\bm A_{n,m}^{pool}\in \mathbb{R}^{B^q \times B^k}$.
Specifically, based on intra-block consistency, we replicate $A_{n,m}^{pool}$ along the query dimension to maintain shared weights within block, while uniformly distributing it along the key dimension to satisfy the softmax constraint, which can be formulated as:
\begin{align}
    \bm A_{n,m}^{pool} = \frac{1}{B^k}\left ( A_{n,m}^{pool} \otimes \mathbf{1}_{B^q \times B^k}\right )
\end{align}  
Here, the Kronecker $\otimes$ expands $A_{n,m}^{pool}$ into the token-level dimensions of $B^{q} \times B^{k}$.

However, the pooling operation only captures the overall block-level weights, failing to reflect the fine-grained variations among tokens within each block.
Therefore, while pooling rectification introduces attention gains, it also incurs pooling errors, and the trade-off between the two leads to unstable rectification performance.

To address this issue, we propose \textbf{Gain-Aware Pooling Rectification (GAPR)}.
Specifically, we theoretically estimate the attention gains and pooling errors at block level.
GAPR performs rectification only when the estimated gain exceeds the error, thereby ensuring a stable performance improvement.
The detailed implementation is as follows:

\noindent \textbf{Attention Gain.}
According to the analysis in Sec~\ref{sec:3.2}, the pooled attention weights exhibit high consistency with the full attention in both magnitude and distribution, providing a reasonable assumption for block-level gain modeling.
Based on this, we define the attention gain $\bm G \in \mathbb{R}^{N \times M}$ of non-critical blocks as the incremental change in attention weights before and after the rectification:
\begin{equation}
\begin{aligned}
    G_{n,m}&=\bm A_{n,m}^{pool} - \bm A_{n,m}^{spa}=\sum_{i\in B_n^q} \sum_{j\in B_m^k} |a_{i,j}^{pool}| \\
    &=\sum_{i\in B_n^q} \sum_{j\in B_m^k} \text{softmax}(s_{i,j}^{pool}) 
\end{aligned}    
\end{equation}
 
\noindent \textbf{Pooling Error.}
To estimate the bias introduced by the pooling approximation, we define the pooling error $\bm E \in \mathbb{R}^{N \times M}$, which quantifies the discrepancy between the real attention weights and the pooled attention weights.
For a non-critical block, the pooling error is formulated as:
\begin{align}
    &E_{n,m} 
    = \bm A_{n,m} - \bm A_{n,m}^{pool}
    = \sum_{i\in B_n^q} \sum_{j\in B_m^k} |a_{i,j} - a_{i,j}^{pool}| \notag \\
    &\overset{(a)}= \sum_{i\in B_n^q} \sum_{j\in B_m^k} |\Delta a_{i,j}|
    = \sum_{i\in B_n^q} \sum_{j\in B_m^k} \text{softmax}(\Delta s_{i,j}) 
\end{align}   
Here, (a) given that the softmax for the pooled and true weights are approximately equivalent, as proven in Appendix.
$\Delta s_{i,j}$ can be efficiently computed as:
\begin{equation}
\begin{aligned}
    \Delta s_{i,j} &= \frac{{q_i} {k_j}^T-q_i^{pool} {k_j^{pool}}^T}{\sqrt{d}} \\
    &\overset{(b)}= \frac{(q_i-q_i^{pool}) {k_j^{pool}}^T + q_i^{pool}{(k_j-k_j^{pool})}^T}{\sqrt{d}}
\end{aligned} 
\end{equation}
where, (b) we omit higher-order error $\frac{{(q_i-q_i^{pool}){(k_j-k_j^{pool})}^T}}{\sqrt{d}}$.

\noindent \textbf{Rectification.}
To ensure stable performance improvement, GAPR performs rectification only when the attention gains exceeds the pooling errors, i.e., $|G_{n,m}| > |E_{n,m}|$.
As shown in Fig.~\ref{fig:overview}, this evaluation is performed at block level, producing a compensation mask $\bm M_c \in \{0,1\}^{N \times M}$, where
$\bm M_c=1$ indicates that the corresponding block weights will be compensated by the implicit full attention:
\begin{align}
    {\bm A^{spa}_{n,m}}' = \bm A_{n,m}^{pool} \approx \bm A_{n,m}
\end{align}  
Here, ${\bm A^{spa}_{n,m}}'$ denotes the rectified sparse attention of the non-critical tokens.
Due to the monotonicity of the softmax function, we relax the rectification condition as:
\begin{align}
    |\sum_{i\in B_n^q} \sum_{j\in B_m^k} s_{i,j}^{pool} | > |\sum_{i\in B_n^q} \sum_{j\in B_m^k} \Delta s_{i,j} |
\end{align}
This eliminates the softmax, further improving efficiency.

\begin{table*}[t]\footnotesize 
    \centering
    \vspace{-0.3cm}
    \caption{Quality and efficiency results on HunyuanVideo-T2V (50 steps, 128 frames, 720p videos).}
    \vspace{-0.3cm}
    \label{tab:hunyuan}
    \setlength{\tabcolsep}{1.8mm}
    
\begin{tabular}{l | c c c c c c c | c c c c}
  \toprule
  \bf Method & \bf VR$\,\uparrow$ & \bf I.Q.$\,\uparrow$ & \bf A.Q.$\,\uparrow$ & \bf S.C.$\,\uparrow$ & \bf M.S.$\,\uparrow$ & \bf B.C.$\,\uparrow$ & \bf VBench$\,\uparrow$ & \bf Sparsity & \bf FLOPs & \bf Latency\,(s) & \bf Speedup \\
  \midrule
  HunyuanVideo & 0.0986 & 67.31 & 56.20 & 96.33 & 99.31 & 96.65 & 83.16 & 0\% & 612.38 PFLOPs & 2425 & 1.00$\times$ \\
  \midrule
  SVG & 0.0890 & 64.01 & 55.84 & 94.64 & 99.35 & 95.88 & 81.94 & 78.91\% & 320.38 PFLOPs & 1010 & 2.43$\times$ \\
  SVG2 & 0.0906 & 64.30 & 54.97 & 94.45 & 99.35 & 95.73 & 81.76 & 79.26\% & 314.27 PFLOPs & 986 & 2.46$\times$ \\
  GRAT-R & 0.0752 & 65.12 & 56.11 & 93.84 & 99.23 & 95.96 & 82.05 & 86.14\% & 289.39 PFLOPs & 830 & 2.92$\times$ \\
  RadialAttention & 0.0819 & 62.80 & \bf 59.17 & 93.54 & 99.31 & 95.79 & 82.12 & 78.25\% & 323.75 PFLOPs & 973 & 2.49$\times$ \\
  Jenga & 0.0855 & \bf 66.39 & 55.74 & 95.01 & 99.12 & 95.90 & 82.43 & 74.64\% & 335.38 PFLOPs & 1050 & 2.31$\times$ \\
  \rowcolor[HTML]{F2F7FB}Ours & \bf 0.0965 & 65.99 & 56.30 & \bf 96.82 & 99.37 & \bf 97.19 & \bf 83.13 & 79.68\% & 310.27 PFLOPs & 970 & 2.50$\times$ \\
  \rowcolor[HTML]{DFECF6}Ours & 0.0890 & 64.32 & 55.92 & 96.10 & \bf 99.42 & 97.08 & 82.57 & \bf 88.95\% & 278.11 PFLOPs & 729 & 3.33$\times$ \\
  \rowcolor[HTML]{C9DFF0}Ours+Tea & 0.0883 & 64.61 & 55.61 & 96.39 & 99.33 & 96.73 & 82.53 & 78.36\% & \bf 180.91 PFLOPs & \bf 463 & \bf 5.24$\times$ \\
  \bottomrule
\end{tabular}
    \vspace{-0.5cm}
\end{table*}

\vspace{-0.1cm}
\begin{algorithm}[th]
    \caption{\quad Overall Process for Rectified SpaAttn}
    \label{alg:algorithm}
    \leftline{\textbf{Require:} $\bm Q$, $\bm K$, $\bm V$, the number of video and text tokens}
    \leftline{$T_v, T_t$, $top$-$k$, weight threshold $p$, adjacency mask $\bm M_{adja}$}
    \leftline{\textbf{Ensure:} Attention output}
    \begin{algorithmic}[1] 
        \STATE Get video and text querys $\bm Q_v, \bm Q_t \gets \bm Q$
        \STATE Get video and text keys $\bm K_v, \bm K_t \gets \bm K$
        \STATE $\bm Q_v^{pool}, \bm K_v^{pool}, \bm V^{pool} \gets$ Pooling($\bm Q_v$), Pooling($\bm K_v$), Pooling($\bm V$), uniform pooling per block.
        \STATE {\color[HTML]{1E90FF}{Implicit full-attention computation:}}
        \STATE \hspace{1.3em}Mix keys: $\bm K_{mix}^{pool} \gets Concat(\bm K_v^{pool}, \bm K_t)$
        \STATE \hspace{1.3em}Mix weight: $\bm A_{mix}^{pool} \gets \text{softmax}(\frac{\bm Q_v^{pool} {\bm K_{mix}^{pool}}^T}{\sqrt{d}})$
        \STATE \hspace{1.3em}Pooled weight: $\bm A^{pool} \gets Reallocate(\bm A_{mix}^{pool})$
        \STATE {\color[HTML]{1E90FF}{Compensation mask generation:}}
        \STATE \hspace{1.3em}Attention gain: $\bm G \gets G_{n,m}=\sum_{i,j\in B_n^q, B_m^k} s_{i,j}^{pool}$
        \STATE \hspace{1.3em}Pooling error: $\bm E \gets E_{n,m}=\sum_{i,j\in B_n^q, B_m^k}\Delta s_{i,j}$
        \STATE \hspace{1.3em}Compensation mask: $\bm M_c \gets |\bm G|>|\bm E|$
        \STATE {\color[HTML]{1E90FF}{Sparse mask generation:}}
        \STATE \hspace{1.3em}$\bm M_I \gets$ $top$-$k(\bm A^{pool})$ $\lor$ $(\sum \bm A^{pool} \odot \bm M_I>p)$ 
        \STATE \hspace{1.3em}Sparse mask: $\widehat{\bm M} \gets \bm M_I \lor \bm M_{adja}$
        \STATE {\color[HTML]{009901}{Block-wise sparse attention:}}
        \STATE \hspace{1.3em}Output: $\bm O_v \gets SparseKernel(\bm Q_v, \bm K, \bm V, \widehat{\bm M})$
        \STATE \hspace{1.3em}Text output: $\bm O_t \gets FullKernel(\bm Q_t, \bm K, \bm V)$
        \STATE {\color[HTML]{CB0000}{Attention bias rectification:}}
        \STATE \hspace{1.3em}Rectified factor:$\bm R \gets R_n=\textstyle\sum_{m=1}^{M} A_{n,m}^{pool} \cdot \widehat{M}_{n,m}$
        \STATE \hspace{1.3em}Rect cri: $\bm O_v^{cri} \gets \bm O_v \odot \bm R$
        \STATE \hspace{1.3em}Rect ncri: $\bm O_v^{ncri} \gets \bm A^{pool} \odot (\neg\widehat{\bm M} \land \bm M_c) \cdot \bm V^{pool}$ 
        \STATE \hspace{1.3em}Rectified output: $\bm O_v' \gets \bm O_v^{cri} + \bm O_v^{ncri}$ 
        \STATE \textbf{Return} $Concat(\bm O_v', \bm O_t)$
    \end{algorithmic}
\end{algorithm}
\vspace{-0.1cm}
\subsection{Implementation and Modularization}
Rectified SpaAttn implements dynamic attention sparsity in three steps: \raisebox{-0.6pt}{\scalebox{1.2}{\ding{172}}} implicit full-attention computation with mask generation, \raisebox{-0.6pt}{\scalebox{1.2}{\ding{173}}} block-wise sparse attention, and \raisebox{-0.6pt}{\scalebox{1.2}{\ding{174}}} attention bias rectification.
Alg.~\ref{alg:algorithm} shows the overall process.
\begin{itemize}

\item IAPR pools multimodal tokens to compute pooled attention weights, which serve as the implicit full attention to assess block importance and guide bias rectification. 
GAPR evaluates attention gains and pooling errors for non-critical blocks, activating compensation mask.
For sparse mask generation, similar to Jenga~\cite{zhang2025training}, we select the $top$-$k$ important blocks with weights above a threshold $p$, while retaining spatial–temporal neighbors blocks.
\item We build sparse attention on FlashAttention2 with Triton, enabling efficient block-wise computation driven by the sparse mask.
Meanwhile, other attention computations remain block-wise to further improve efficiency.
\item Bias rectification is applied to the sparse attention results.
For critical tokens, weighted correction is performed using rectification factors derived from the implicit full attention.
For non-critical tokens, attention information is recovered via the compensation mask by combining the implicit full attention with pooled values.
\end{itemize}

\begin{figure}[!h]
    \centering
    \vspace{-0.3cm}
    \includegraphics[width=0.48\textwidth]{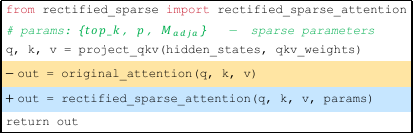}
    \vspace{-0.5cm}
    \caption{Minimal Integration of Rectified SpaAttn.}
    \label{fig:code}
    \vspace{-0.5cm}
\end{figure}

Since all rectification are built on pooled vectors, the additional overhead is negligible.
To improve usability and transferability, Rectified SpaAttn is encapsulated as a plug-and-play module, enabling seamless integration into other DiT models with minimal code, as shown in Fig.~\ref{fig:code}.

\section{Experiment}
\vspace{-0.1cm}
\noindent \textbf{Models.} We evaluate Rectified SpaAttn on open-sourced video generation models including HunyuanVideo-T2V-13B~\cite{kong2024hunyuanvideo} and Wan2.1-I2V/\,T2V-14B~\cite{wan2025wan}. We further extend it to image generation model Flux.1-dev-12B~\cite{blackforest2024flux}.

\noindent \textbf{Metrics.} 
We use \textit{Vision Reward} (VR)~\cite{xu2024visionreward}, and VBench~\cite{huang2024vbench} key metrics to evaluate video quality, including {\textit{imaging quality} (I.Q.), 
\textit{aesthetic quality} (A.Q.), 
\textit{subject consistency} (S.C.), 
\textit{motion smoothness} (M.S.), and \textit{background consistency} (B.C.)}. 
Their average is reported as VBench.
We evaluate image quality using \textit{Image Reward} (IR$\,\uparrow$)~\cite{xu2023imagereward}, \textit{FID}$\,\downarrow$~\cite{heusel2017gans}, \textit{CLIP}$\,\uparrow$~\cite{radford2021learning}, 
\textit{SSIM}$\,\uparrow$, 
\textit{PSNR}$\,\uparrow$, and \textit{LPIPS}$\,\downarrow$.
The efficiency of the sparsity mechanism is quantified by the \textit{Sparsity}, defined as the computations removed relative to full attention.
To intuitively evaluate the end-to-end efficiency, we further measure \textit{FLOPs}, \textit{Latency}, and \textit{Speedup}.

\begin{table}[t]\scriptsize 
    \centering
    \caption{Generation results on Flux.1-dev (50 steps, 4K images).}
    \vspace{-0.3cm}
    \label{tab:flux}
    \setlength{\tabcolsep}{0.95mm}
    

\begin{tabular}{l | c c c c c c | c c}
  \toprule
  \bf Method & \bf IR & \bf FID & \bf CLIP & \bf SSIM & \bf PSNR & \bf LPIPS & \bf Sparsity & \bf Speedup \\
  \midrule
  Flux.1-dev & 1.0760 & - & 26.50 & - & - & - & 0\% & 1.00$\times$ \\
  \midrule
  GRAT-R & 0.8415 & 15.59 & 26.34 & 0.734 & 19.58 & 0.283 & \bf 96.20\% & 1.27$\times$ \\
  Jenga & 0.9562 & 12.31 & 26.35 & 0.953 & 32.10 & 0.032 & 73.86\% & 1.32$\times$ \\
  \rowcolor[HTML]{F2F7FB}Ours & \bf 1.0131 & \bf 5.37 & 26.43 & \bf 0.976 & \bf 36.89 & \bf 0.015 & 74.05\% & 1.40$\times$ \\
  \rowcolor[HTML]{DFECF6}Ours & 1.0050 & 10.15 & \bf 26.48 & 0.965 & 33.26 & 0.029 & 87.66\% & 1.60$\times$ \\
  \rowcolor[HTML]{C9DFF0}Ours+Tea & 1.0047 & 10.23 & 26.47 & 0.964 & 33.26 & 0.029 & 87.66\% & \bf 4.15$\times$ \\
  \bottomrule
\end{tabular}
    \vspace{-0.6cm}
\end{table}

\begin{table*}[t]\footnotesize 
    \centering
    \vspace{-0.3cm}
    \caption{Quality and efficiency results on Wan2.1-T2V / I2V (50 steps, 81 frames, 720p videos).}
    \vspace{-0.3cm}
    \label{tab:wan}
    \setlength{\tabcolsep}{1.8mm}
    
\begin{tabular}{l | c c c c c c c | c c c c}
  \toprule
  \bf Method & \bf VR$\,\uparrow$ & \bf I.Q.$\,\uparrow$ & \bf A.Q.$\,\uparrow$ & \bf S.C.$\,\uparrow$ & \bf M.S.$\,\uparrow$ & \bf B.C.$\,\uparrow$ & \bf VBench$\,\uparrow$ & \bf Sparsity & \bf FLOPs & \bf Latency\,(s) & \bf Speedup \\
  \midrule
  Wan2.1-T2V & 0.1029 & 68.82 & 60.56 & 96.24 & 98.68 & 96.43 & 84.15 & 0\% & 658.46 PFLOPs & 2731 & 1.00$\times$ \\
  \midrule
  SVG & 0.0863 & 65.24 & \bf 60.66 & 95.84 & 98.74 & 96.28 & 83.35 & 75.90\% & 413.11 PFLOPs & 1631 & 1.67$\times$ \\
  SVG2 & 0.0979 & 65.58 & 60.09 & 96.32 & 99.04 & 96.32 & 83.47 & 75.57\% & 416.57 PFLOPs & 1643 & 1.66$\times$ \\
  RadialAttention & 0.0977 & 65.09 & 60.05 & 96.29 & 99.19 & 96.72 & 83.46 & 75.91\% & 414.25 PFLOPs & 1634 & 1.67$\times$ \\
  Jenga & 0.1005 & 65.46 & 60.14 & 95.96 & 98.85 & 96.42 & 83.37 & 73.69\% & 419.88 PFLOPs & 1656 & 1.65$\times$ \\
  \rowcolor[HTML]{F2F7FB}Ours & \bf 0.1026 & \bf 65.67 & 59.78 & 96.91 & \bf 99.21 & 97.01 & \bf 83.72 & 74.88\% & 415.34 PFLOPs & 1624 & 1.68$\times$ \\
  \rowcolor[HTML]{DFECF6}Ours & 0.0991 & 65.40 & 58.94 & 96.87 & 99.20 & 96.92 & 83.47 & \bf 79.71\% & 398.75 PFLOPs & 1515 & 1.80$\times$ \\
  \rowcolor[HTML]{C9DFF0}Ours+Tea & 0.0903 & 63.68 & 58.86 & \bf 97.03 & 99.18 & \bf 97.08 & 83.17 & 74.82\% & \bf 173.46 PFLOPs & \bf 592 & \bf 4.61$\times$ \\
  \midrule
  Wan2.1-I2V & 0.1405 & 71.41 & 60.39 & 93.14 & 98.02 & 94.68 & 83.53 & 0\% & 645.45 PFLOPs & 2754 & 1.00$\times$ \\
  \midrule
  SVG & 0.1156 & 71.07 & 59.23 & 91.99 & 97.94 & 93.73 & 82.79 & 69.27\% & 428.75 PFLOPs & 1794 & 1.54$\times$ \\
  SVG2 & 0.1256 & 71.54 & 59.93 & 92.96 & 97.97 & 94.38 & 83.35 & 70.11\% & 425.64 PFLOPs & 1763 & 1.56$\times$ \\
  Jenga & 0.1142 & 70.02 & 57.90 & 90.88 & 97.86 & 92.92 & 81.92 & 73.68\% & 385.27 PFLOPs & 1543 & 1.78$\times$ \\
  \rowcolor[HTML]{F2F7FB}Ours & \bf 0.1391 & \bf 71.79 & \bf 60.65 & 93.70 & 98.48 & 94.84 & \bf 83.89 & 74.69\% & 375.17 PFLOPs & 1522 & 1.81$\times$ \\
  \rowcolor[HTML]{DFECF6}Ours & 0.1288 & 71.09 & 59.71 & 93.29 & 98.53 & 94.52 & 83.43 & \bf 83.91\% & 348.56 PFLOPs & 1327 & 2.08$\times$ \\
  \rowcolor[HTML]{C9DFF0}Ours+Tea & 0.1195 & 70.85 & 59.62 & \bf 94.34 & \bf 99.03 & \bf 95.51 & 83.87 & 74.64\% & \bf 91.34 PFLOPs  & \bf 307 & \bf 8.97$\times$ \\
  \bottomrule
\end{tabular}
    \vspace{-0.5cm}
\end{table*}

\noindent \textbf{Datasets.} 
We use prompts in Penguin Benchmark~\cite{kong2024hunyuanvideo} for text-to-video generation, using prompt-image pairs from VBench~\cite{huang2024vbench} and cropping images to match 720p resolution for image-to-video generation, and using the same COCO prompts as GRAT~\cite{ren2025grouping} for text-to-image generation.

\noindent \textbf{Baselines.} 
We compare Rectified SpaAttn with state-of-the-art sparse methods including SVG~\cite{xi2025sparse}, SVG2~\cite{yang2025sparse}, GRAT-R~\cite{ren2025grouping}, RadialAttention~\cite{li2025radial}, and Jenga~\cite{zhang2025training}. 
To ensure a fair comparison, we match their efficiency to ours by tuning the sparsity ratio, as detailed in the Appendix.

\noindent \textbf{Implementation Details.} 
We implement Rectified SpaAttn as a plug-and-play module with customized kernels based on FlashAttention~\cite{dao2023flashattention}, and benchmark it on an NVIDIA H100 PCIe with CUDA 12.4. Rectified SpaAttn introduces no empirical parameters, allowing users to flexibly balance efficiency and quality by adjusting the sparsity ratio. Details of the experimental setup are provided in the Appendix. For Wan2.1-T2V, we retain a 5-step warm-up following prior work~\cite{li2025radial,yang2025sparse}.
We further integrate it with the caching method TeaCache~\cite{liu2025timestep} to achieve a higher speedup.

\begin{table}[t]\footnotesize 
    \centering
    \caption{Ablation study of IPAR and GAPR on HunyuanVideo.}
    \vspace{-0.3cm}
    \label{tab:abl}
    \setlength{\tabcolsep}{1.2mm}
    \begin{tabular}{ccc|cc|cc}
    \toprule 
    \multicolumn{3}{c|}{\bf Method} & \multirow{2}{*}{\centering \bf VR$\,\uparrow$} & \multirow{2}{*}{\centering \bf VBench$\,\uparrow$} & \multirow{2}{*}{\centering \bf Latency} & \multirow{2}{*}{\centering \bf Speedup} \\ 
    \bf Rectify & \bf IPAR & \bf GAPR & & & & \\ 
    \midrule
    \xmark & \xmark & \xmark & 0.0585 & 81.15 & 719 s & 3.37$\times$ \\ 
    \cmark & \xmark & \xmark & 0.0435 & 79.85 & 725 s & 3.34$\times$ \\ 
    \cmark & \cmark & \xmark & 0.0805 & 81.92 & 726 s & 3.34$\times$ \\ 
    \cmark & \cmark & \cmark & 0.0890 & 82.57 & 729 s & 3.33$\times$ \\ 
    \bottomrule
\end{tabular}
    \vspace{-0.5cm}
\end{table}

\subsection{Quality and Efficiency Evaluation}
\vspace{-0.1cm}
As reported in Tab.~\ref{tab:hunyuan},~\ref{tab:flux}, and~\ref{tab:wan}, at comparable sparsity, Rectified SpaAttn consistently outperforms all baselines while achieving higher speedups.
As the sparsity increases, Rectified SpaAttn not only pushes the limits of sparse speedup but also maintains high quality.
Moreover, it can be seamlessly combined with caching techniques to further boost inference efficiency.
Specifically, Rectified SpaAttn achieves VR scores of 0.1026 and 0.1391 on Wan2.1-T2V and Wan2.1-I2V, respectively, at approximately 75\% sparsity. On HunyuanVideo, it attains a 3.33$\times$ speedup at 88.95\% sparsity while achieving a VBench Score of 82.57.
By combining with TeaCache, our method achieves 5.24$\times$, 8.97$\times$, and 4.15$\times$ speedups on HunyuanVideo-T2V, Wan2.1-I2V, and Flux.1-dev, respectively.

\subsection{Ablation Study}
\vspace{-0.1cm}
\textbf{Effectiveness and Efficiency of IPAR and GAPR.}
We use Jenga with 88.95\% sparsity as the baseline.
As shown in Tab.~\ref{tab:abl}, under high sparsity, Jenga’s performance degrades significantly, with the VR score dropping to 0.0585.
When rectification is applied using directly pooled attention map, performance further degrades, indicating its poor alignment with full attention, leading to misleading rectifications.
By introducing IPAR, the performance improves significantly, indicating that IPAR effectively captures distribution-aligned implicit full attention, thereby rectifying the bias of critical tokens.
GAPR effectively constrains the rectification of non-critical tokens to ensure the gains surpass pooling errors, thereby achieving stable performance improvements.
Moreover, they incur negligible overhead with minimal impact on inference efficiency.

\begin{figure}[!t]
    \centering
    \includegraphics[width=0.42\textwidth]{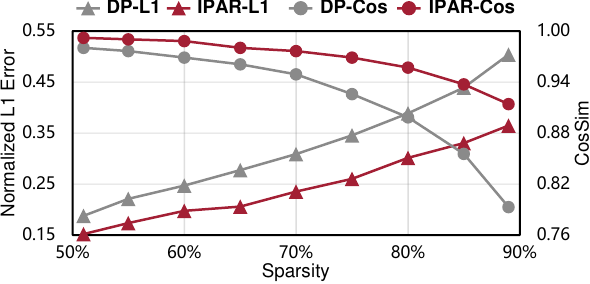}
    \vspace{-0.4cm}
    \caption{Alignment evaluation of IPAR on HunyuanVideo.}
    \label{fig:ab_ipar}
    \vspace{-0.3cm}
\end{figure}

\noindent \textbf{Implicit Full Attention Alignment.}
Alignment between the implicit and the full attention serves as the foundation for effective rectification.
We evaluate this alignment both numerically and distributionally using normalized L1 error and cosine similarity, respectively.
As shown in Fig.~\ref{fig:ab_ipar}, compared with directly pooled attention weights (DP), the implicit full attention obtained by IPAR exhibits higher consistency with the true full attention, demonstrating its effectiveness in attention alignment.

\begin{figure}[!t]
    \centering
    \includegraphics[width=0.42\textwidth]{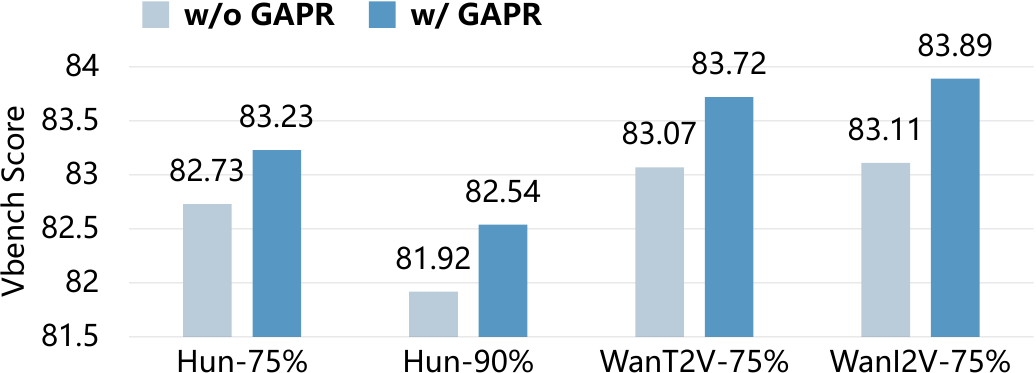}
    \vspace{-0.2cm}
    \caption{Effectiveness evaluation of GAPR.}
    \label{tab:ab_gapr}
    \vspace{-0.7cm}
\end{figure}
\noindent \textbf{Advantages of Gain-Aware Rectification.}
We validate the effectiveness of GAPR across different models and sparsity levels. As shown in Fig.~\ref{tab:ab_gapr}, GAPR consistently and significantly improves the performance of sparse models.

\vspace{-0.1cm}
\section{Conclusion}
\vspace{-0.1cm}
In this paper, we revisit attention sparsity and propose a new paradigm that improves sparse attention accuracy through bias rectification.
Building on this, we propose Rectified SpaAttn, which rectifies attention allocation biases with implicit full attention reference, thereby enhancing the alignment between sparse and full attention maps.
By the isolated-pooling attention reallocation, Rectified SpaAttn derives a distribution-aligned implicit full attention, effectively rectifying the attention bias of critical tokens.
By the gain-aware pooling rectification, Rectified SpaAttn ensures that the rectified gains consistently surpass the pooling errors, effectively rectifying the attention bias of non-critical tokens.
Extensive experiments show that Rectified SpaAttn achieves superior generation quality even under high sparsity, making video generation more efficient and practical.

{
    \small
    \bibliographystyle{ieeenat_fullname}
    \bibliography{main}
}


\clearpage
\setcounter{page}{1}
\maketitlesupplementary

\section{Proof: Equivalence Analysis of Softmax Normalization under Pooling for DiTs}\label{proof}
In this section, we show that the pooled attention weights of DiT visual tokens share approximately equivalent softmax normalization denominators with the true attention weights, allowing the $\Delta w_{i,j} = w_{i,j} - w_{i,j}^{pool}$.
The proof proceeds as follows.

For query $q_i$, the softmax denominator of $w_{i,j}$ is expressed as:
\begin{equation}
\begin{aligned}\label{eq:Ssumi}
    S_{sum}(i)
    &=\sum_{m=1}^{M} \sum_{j\in B_m^k} \text{exp}(\frac{q_i k_j^T}{\sqrt{D}}) \\
    &=\sum_{m=1}^{M} \sum_{j\in B_m^k} \text{exp}(s_{i,j}) 
\end{aligned}  
\end{equation}
The softmax denominator of $w_{i,j}^{pool}$ is expressed as:
\begin{equation}
\begin{aligned}\label{eq:Spoolsumi}
    S_{sum}^{pool}(i)
    &=\sum_{m=1}^{M} \sum_{j\in B_m^k} \text{exp}(\frac{q_i^{pool} {k_j^{pool}}^T}{\sqrt{D}}) \\
    &=\sum_{m=1}^{M} \sum_{j\in B_m^k} \text{exp}(s_{i,j}^{pool})
\end{aligned}
\end{equation}
For $q_i$ interacting with the m-th key block $K_m$, the average intra-block attention score $s_{i,m}$ is given by:
\begin{align}
    s_{i,m}=\frac{1}{B^k} \sum_{j\in B_m^k} s_{i,j} = \frac{q_i {k_j^{pool}}^T}{\sqrt{D}}
\end{align}
The softmax denominator of $w_{i,j}$ within this block is:
\begin{align} \label{eq:Smi}
    S_{m}(i)
    &=\sum_{j\in B_m^k}^{} \text{exp}(s_{i,j}) \notag \\
    &= B^k \cdot \text{exp}(s_{i,m}) \cdot \frac{1}{B^k} \sum_{j\in B_m^k} \text{exp}(s_{i,j}-s_{i,m}) \notag \\
    &= B^k \cdot \text{exp}(s_{i,m}) \cdot A_{i,m}
\end{align}
We define:
\begin{align}
    A_{i,m} = \frac{1}{B^k} \sum_{j \in B_m^k} \exp(s_{i,j} - s_{i,m}) = \mathbb{E}[\exp(\varepsilon_{i,j})]
\end{align}
where $\varepsilon_{i,j} = s_{i,j} - s_{i,m}$ denotes the zero-mean intra-block deviation of the attention scores.
Applying a second-order Taylor expansion to the exponential function with respect to $\varepsilon$:
\begin{align}
    \text{exp}(\varepsilon_{i,j})=1+\varepsilon+\frac{1}{2} \varepsilon^2+O(\varepsilon^3)
\end{align}
Given the randomness of the pooling error, $\mathbb{E}[\varepsilon_{i,j}] = 0$, and omitting higher-order terms, $A_{i,m}$ simplifies to:
\begin{align}\label{eq:A_im}
    A_{i,m}=1+\frac{1}{2} \varepsilon^2
\end{align}
Here, $\varepsilon^2$ essentially corresponds to the variance of the true intra-block attention scores.
Substituting Eq.~\ref{eq:Smi} and Eq.~\ref{eq:A_im} into Eq.~\ref{eq:Ssumi} yields:
\begin{align} \label{eq:Srealsum1}
    S_{sum}(i)=\sum_{m=1}^{M} B^k \cdot \text{exp}(s_{i,m}) \cdot (1+\frac{1}{2} \varepsilon^2)
\end{align}
Previous studies~\cite{blanchard2021accurately,wei2023convex,joulin2017efficient} have shown that introducing small perturbations during softmax computation does not affect its validity.
We define the condition $|S_{\text{sum}} - S_{\text{sum}}^{\text{pool}}| < 5\% |S_{\text{sum}}|$ as a criterion for introducing small perturbations.
Based on Eq.~\ref{eq:Spoolsumi} and Eq.~\ref{eq:Srealsum1}, this condition is evaluated across all layers of the HunyuanVideo in a block-wise manner.
Experimental statistics show that 99.3\% of block interactions satisfy this criterion, thereby demonstrating the approximate equivalence of the softmax normalization denominators between the true and pooled attention weights.

\begin{figure*}[!t]
    \centering
    \includegraphics[width=0.90\textwidth]{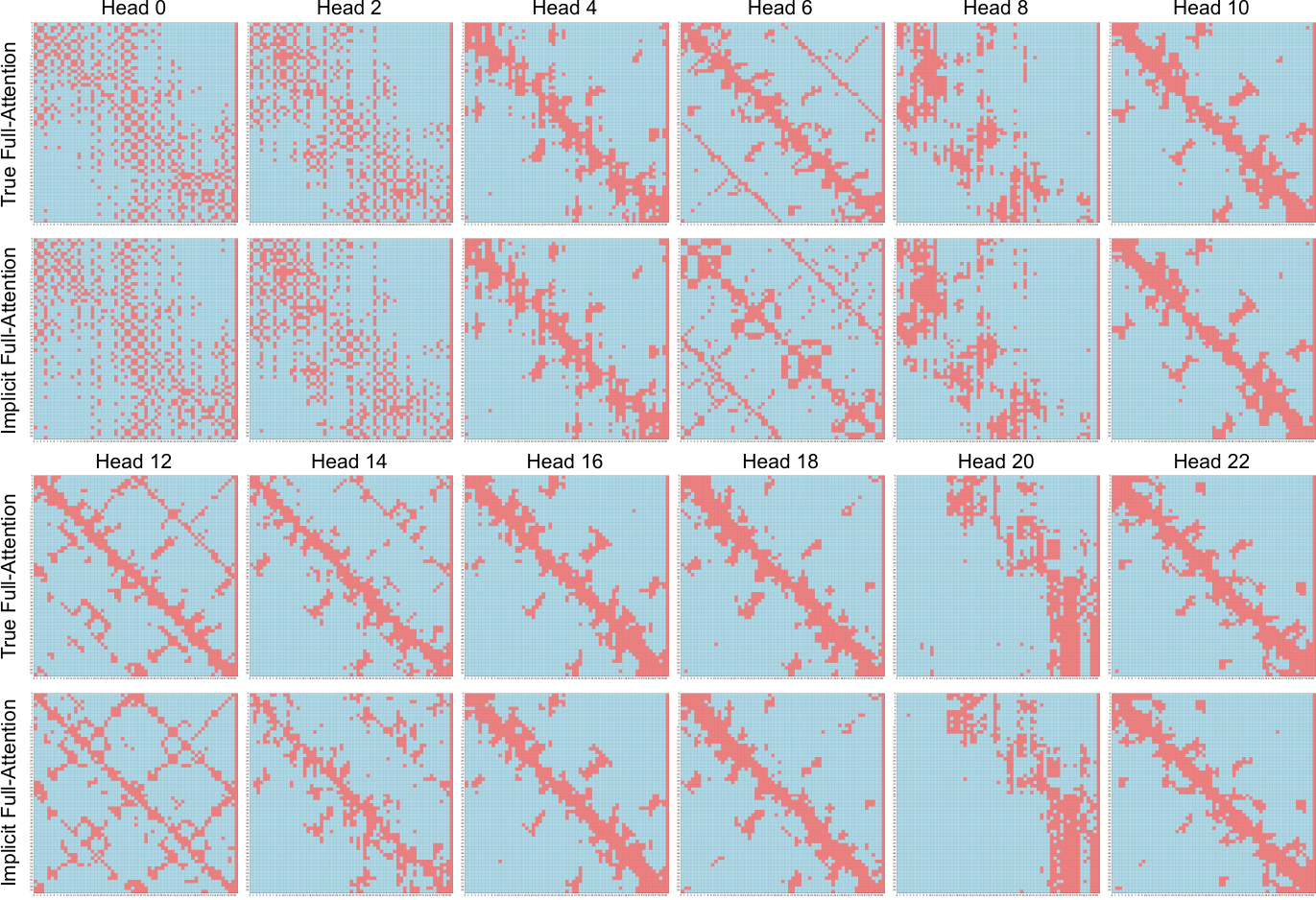}
    \caption{Implicit and truth full-attention maps of the first attention layer in HunyuanVideo under a 7,680 token sequence.}
    \label{fig:ipar_visualization}
    \vspace{-0.4cm}
\end{figure*}

\section{Implementations}

\paragraph{Baselines.}
We compare our method with state-of-the-art baselines including SVG~\cite{xi2025sparse}, SVG2~\cite{yang2025sparse}, GRAT~\cite{ren2025grouping}, RadialAttention~\cite{li2025radial}, and Jenga~\cite{zhang2025training}. 
We evaluate performance by directly using the official open-source implementations. For a fair comparison, we tune their sparsity parameters to match the efficiency of our method. For SVG2, simply adjusting the sparsity ratio leads to performance degradation; therefore, we appropriately modify its warmup schedule.

\noindent \textbf{SVG:} For HunyuanVideo, Wan2.1-T2V, and Wan2.1-I2V, we set the sparsity parameters  $sparsity$ to 0.15, 0.25, and 0.2, respectively.

\noindent \textbf{SVG2:} For HunyuanVideo, Wan2.1T2V, and Wan2.1I2V, we set the sparsity parameters $top\_p\_k$ parameters to 0.6, 0.7, and 0.75, respectively. 
In addition, the 10-step warmup for HunyuanVideo is reduced to 5 steps.

\noindent \textbf{GRAT:} For HunyuanVideo and Flux.1-dev, we employ local attention windows over surrounding blocks, controlling the window size to 5 in both 3D and 2D attention spaces.

\noindent \textbf{RadialAttention:} For HunyuanVideo and Wan2.1-T2V, we set the $decay$ parameters to 0.6 and 0.2, respectively.

\noindent \textbf{Jenga:} For HunyuanVideo, we set the sparsity parameters $top\_k$ to 0.75 and $p$ to 0.3; for Wan2.1-T2V and Wan2.1-I2V, we set $top\_k$ to 0.8 and $p$ to 0.8; for Flux.1-dev, we set $top\_k$ to 0.5 and $p$ to 0.5.

\paragraph{Rectified SpaAttn.}

Our method involves two parameters for controlling sparsity ratio. $top$-$k$ determines the number of important blocks to retain, while $p$ sets the threshold such that the sum of attention weights of the retained blocks exceeds this value.
Prior to inference, we reorder the token sequence using a Space-Filling Curve~\cite{zhang2025training} to enhance intra-block consistency. Except for retaining a 5-step warmup for Wan2.1-T2V, no warmup is applied to other models.
When combined with Teacache~\cite{liu2025timestep}, we preserve its existing cache architecture and do not modify any cache parameters.
The settings for the main experimental results are summarized in Table~\ref{tab:param}.
The table colors are aligned with those used in the main results.

\section{Visualization of Generation Results}
We provide the videos and images generated by Rectified SpaAttn on our open-source repository at \href{https://github.com/BienLuky/Rectified-SpaAttn}{https://github.com/BienLuky/Rectified-SpaAttn}.

\begin{table}[t]
    \centering
    \caption{Detailed parameters of the main results. }
    \vspace{-0.2cm}
    \label{tab:param}
    \resizebox{0.45\textwidth}{!}{
    \begin{tabular}{c|c|ccc}
\toprule
\bf Model & \bf Settings & \bf $top$-$k$ & \bf $p$ & \bf $tea$-$thresh$ \\ 
\midrule

\rowcolor[HTML]{F2F7FB} 
\cellcolor[HTML]{FFFFFF}& Ours & 0.2 & 0.3 & - \\  

\rowcolor[HTML]{DFECF6} 
\cellcolor[HTML]{FFFFFF}& Ours & 0.1 & 0.3 & - \\  

\rowcolor[HTML]{C9DFF0} 
\cellcolor[HTML]{FFFFFF}\multirow{-3}{*}{HunyuanVideo} & Ours+Tea & 0.2 & 0.3 & 0.15 \\
\midrule

\rowcolor[HTML]{F2F7FB}
\cellcolor[HTML]{FFFFFF}& Ours & 0.25 & 0.3 & - \\  

\rowcolor[HTML]{DFECF6}
\cellcolor[HTML]{FFFFFF}& Ours & 0.2 & 0.3 & - \\  

\rowcolor[HTML]{C9DFF0}
\cellcolor[HTML]{FFFFFF}\multirow{-3}{*}{Wan2.1-T2V} & Ours+Tea & 0.25 & 0.3 & 0.2 \\
\midrule

\rowcolor[HTML]{F2F7FB}
\cellcolor[HTML]{FFFFFF}& Ours & 0.25 & 0.3 & - \\  

\rowcolor[HTML]{DFECF6}
\cellcolor[HTML]{FFFFFF}& Ours & 0.15 & 0.3 & - \\  

\rowcolor[HTML]{C9DFF0}
\cellcolor[HTML]{FFFFFF}\multirow{-3}{*}{Wan2.1-I2V} & Ours+Tea & 0.25 & 0.3 & 0.3 \\
\midrule

\rowcolor[HTML]{F2F7FB}
\cellcolor[HTML]{FFFFFF}& Ours & 0.25 & 0.3 & - \\  

\rowcolor[HTML]{DFECF6}
\cellcolor[HTML]{FFFFFF}& Ours & 0.1 & 0.3 & - \\  

\rowcolor[HTML]{C9DFF0}
\cellcolor[HTML]{FFFFFF}\multirow{-3}{*}{Flux.1-dev} & Ours+Tea & 0.1 & 0.3 & 0.8 \\
\bottomrule
\end{tabular}

    }
    \vspace{-0.3cm}
\end{table}

\section{Visualization of Implicit Full-Attention Maps}

In this section, we visualize the implicit full-attention maps obtained by IPAR. 
As shown in Fig.~\ref{fig:ipar_visualization}, our implicit full-attention exhibits a high degree of distributional alignment with the truth full attention, providing a solid foundation for the effectiveness of our rectification mechanism.

\end{document}